%% file: tooleval.tex
\pdfoutput=1

\documentclass[11pt]{article}
\usepackage[table]{xcolor}
\usepackage{ACL2023}

\usepackage{times}
\usepackage{booktabs}
\usepackage{latexsym}
\usepackage{graphicx}
\usepackage{multirow}
\usepackage{multicol}
\usepackage{hyperref}
\usepackage{listings}
\usepackage{amsmath}
\usepackage[most]{tcolorbox}
\usepackage{float}
\usepackage{xspace}

\usepackage[T1]{fontenc}

\usepackage[utf8]{inputenc}

\newcommand{\shortname}{\texttt{T-Eval}}
\newcommand{\eg}{\textit{e.g.}\@\xspace}

\usepackage{microtype}

\usepackage{inconsolata}

\title{\shortname: Evaluating the Tool Utilization Capability of\\ Large Language Models Step by Step}

\author{Zehui Chen$^{1,2*}$ \quad Weihua Du$^{3,2*}$ \quad Wenwei Zhang$^{2}\thanks{\quad Equal Contributions}$ \quad Kuikun Liu$^{2}$ \quad Jiangning Liu$^{2}$\\
\textbf{Miao Zheng$^{2}$ \quad Jingming Zhuo$^{4,2}$ \quad Songyang Zhang$^{2}$ \quad Dahua Lin$^{2}$ \quad Kai Chen$^{2 \dag}$ \quad Feng Zhao$^{1}\thanks{\quad Corresponding author}$}\\
  $^{1}$University of Science and Technology of China  \quad $^{2}$Shanghai AI Laboratory\\
  $^{3}$Tsinghua University \quad $^{4}$Jilin University\\
}

\begin{document}
\maketitle

\input{sections/0_abstract.tex}
\input{sections/1_intro.tex}
\input{sections/2_method.tex}
\input{sections/3_experiments.tex}
\input{sections/4_related_work.tex}
\input{sections/5_conclusion.tex}

\bibliography{anthology}
\bibliographystyle{acl_natbib}

\appendix
\input{sections/99_appendix.tex}
%

\end{document}

%% file: sections/0_abstract.tex
\begin{abstract}
Large language models (LLMs) have achieved remarkable performance on various NLP tasks and are augmented by tools for broader applications. Yet, how to evaluate and analyze the tool utilization capability of LLMs is still under-explored. In contrast to previous works that evaluate models holistically, we comprehensively decompose the tool utilization into multiple sub-processes, including instruction following, planning, reasoning, retrieval, understanding, and review. Based on that, we further introduce \shortname~to evaluate the tool-utilization capability step by step. \shortname~disentangles the tool utilization evaluation into several sub-domains along model capabilities, facilitating the inner understanding of both holistic and isolated competency of LLMs. We conduct extensive experiments on \shortname~and in-depth analysis of various LLMs. \shortname~ not only exhibits consistency with the outcome-oriented evaluation but also provides a more fine-grained analysis of the capabilities of LLMs, providing a new perspective in LLM evaluation on tool-utilization ability. The benchmark will be available at \href{https://github.com/open-compass/T-Eval}{https://github.com/open-compass/T-Eval}.
\end{abstract}

%% file: sections/1_intro.tex

\begin{figure*}[!t]
    \vspace{-4mm}
    \centering    \includegraphics[width=2.05\columnwidth]{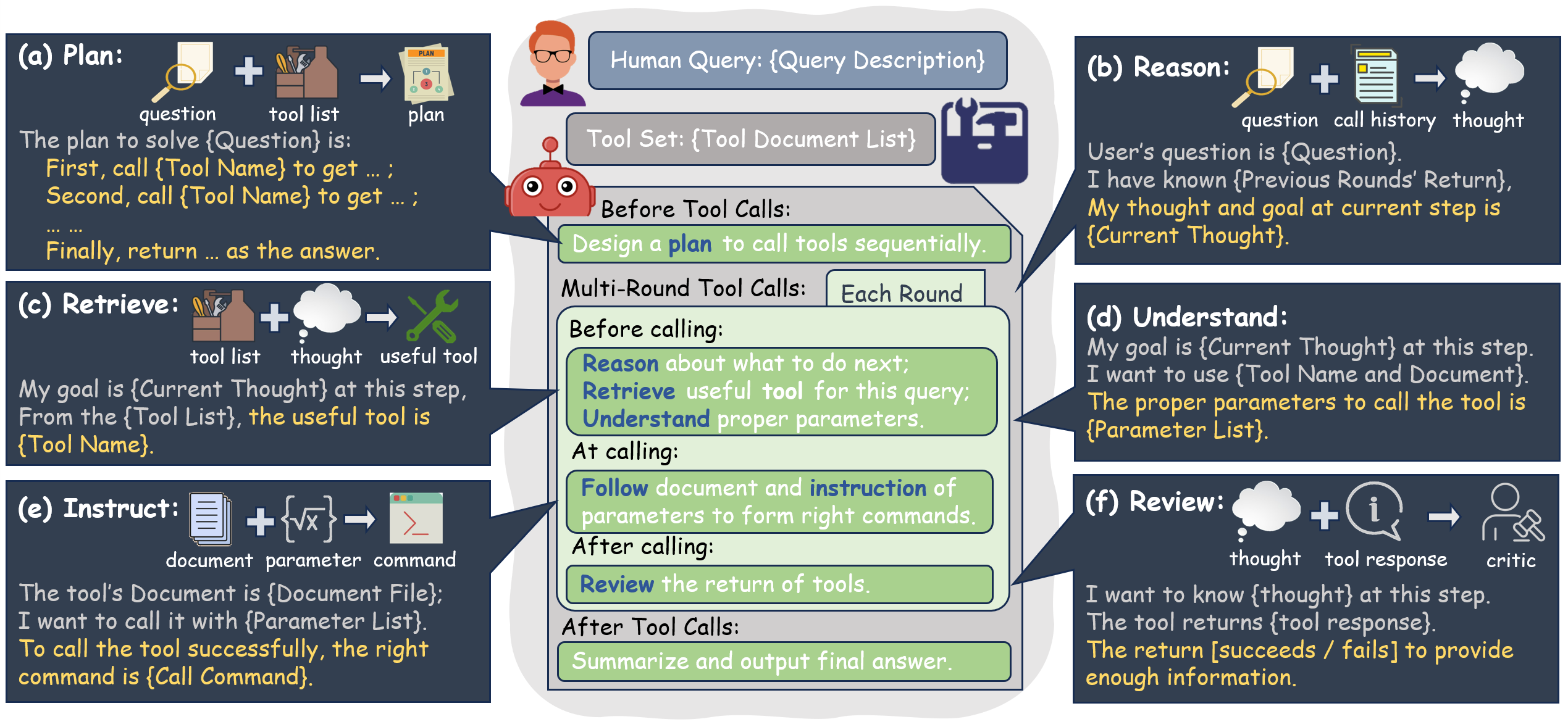}
    \caption{\textbf{Overview of \shortname}: \shortname\ decomposes the tool utilization capability into six necessary abilities: \textsc{plan, reason, retrieve, understand, instruct} and \textsc{review}. To respond to a query with a given tool list, LLM agents generate a \textbf{plan} first before calling tools. The solution path is multiple rounds of tool calling where agents \textbf{reason} their thoughts, \textbf{retrieve} and \textbf{understand} the necessary tools and parameters, execute the \textbf{instructions}, and finally \textbf{review} the tool response.}
    \label{fig: intro}
    \vspace{-2.5mm}
\end{figure*}

\section{Introduction}
Large language models (LLMs) have fueled dramatic progress and emerged as a promising path to more advanced intelligence~\cite{zhao2023survey, kaddour2023challenges}.
To further extend the capability of LLMs, tool utilization, which empowers LLMs to leverage external tools to solve more complicated problems, has spurred vast research interests in both research and industry \cite{parisi2022talm, schick2023toolformer, mialon2023augmented}.

Despite the attractive ability achieved by aiming LLMs with tools, how to evaluate LLMs in tool learning has not been fully explored. Existing works evaluate the tool utilization ability based on the final output~\cite{qin2023toolllm} or only consider the single-step tool calling~\cite{li2023api}. However, real-world problems usually involve complex planning and executing multiple tools. Simply judging the quality through the final output omits the assessment of the intermediate steps, making it hard to identify the main bottlenecks of the tool-use capability in LLMs.
Besides, current benchmarks mainly rely on real-time tool interactions~\cite{qin2023toolllm, li2023api}, which overlooks the external factors (instability of API service or temporal information shift) on the overall judgment, leading to evaluation variance and unfair comparison.


To overcome the above problems, we introduce \shortname, a step-by-step \textbf{T}ool \textbf{Eval}uation benchmark for LLMs. Unlike prior works that appraise the model from a holistic perspective, we explicitly decompose the evaluation into several sub-tasks along the basic capabilities of the language model. Specifically, given the golden tool-utilization annotations verified by human experts, we dedicatedly designed the evaluation protocols and corresponding instruction prompts based on the intermediate steps along the annotation path. Such a paradigm enables us to separately benchmark each competence of the LLMs, including planning, reasoning, retrieval, understanding, instruction following, and review. Additionally, thanks to the decomposed evaluation protocols, our benchmark significantly alleviates the exogenous influences (such as online tools) during the evaluation process, yielding a more stable and fair model assessment.

By conducting extensive experiments on \shortname, we carry out in-depth analysis and insights on the results, pinpointing the main bottlenecks of current LLMs in tool learning. Furthermore, we also prove that our benchmark reveals consistent evaluation of individual and comprehensive model abilities, where higher individual ability scores lead to better performance on complex downstream tasks, providing a new perspective in LLM evaluation on tool utilization. Our major contributions are as follows: 
\begin{itemize}
    \item We introduce \shortname, a step-by-step tool utilization evaluation benchmark, which decomposes the evaluation into several sub-tasks, gauging the fine-grained abilities of LLMs as tool agents.
    \item \shortname~ uses a multi-agent data generation pipeline verified by human experts. This approach significantly reduces the impact of external factors, leading to a more stable and fair assessment of the LLMs.
    \item Extensive experiments conducted with various LLMs validate the effectiveness and generalization of \shortname, providing valuable insights into bottlenecks of current LLMs, and offering new perspectives in improving tool-utilization capabilities.
\end{itemize}

%% file: sections/2_method.tex
\section{T-Eval}
Benchmarking LLMs as tool agents involves multiple dimensions of evaluations of LLM abilities and suffers from the external influence of tools.
Therefore, 
we first thoroughly investigate each critical dimension of the tool-calling process (\S \ref{sec:eval_paradigm}), and then establish tailored evaluation protocols for each dimension (\S \ref{sec:eval_protocol}), named \shortname, to enable a detailed evaluation of tool utilization capability. To guarantee the high quality of the golden solution paths and tool-calling responses, we adopt a human-in-the-loop data generation pipeline (\S \ref{sec:data_construct}), ensuring the stability and longevity of \shortname.
Lastly, we provide a statistical overview of \shortname~(\S \ref{sec:data_summary}).

\subsection{Evaluation Decomposition}
\label{sec:eval_paradigm}

Tool utilization with large language models (LLMs) encompasses a variety of scenarios, touching upon multiple dimensions of capabilities. To better understand the whole process, we first deconstruct the tool-calling process into several key aspects, as depicted in Fig. \ref{fig: intro}.

First, solving complex real-world problems frequently requires a multi-step approach to tool calling. For example, to know the weather from a month ago, an LLM must first confirm the current date before it can query a weather tool. This requires a robust \textbf{planning} ability (Fig. \ref{fig: intro}(a)) to develop a strategy for tool calling that guides subsequent actions.
Moreover, the contexts in which tools are utilized can be intricate, involving tool descriptions, documentation, user queries, previous interactions, and prior observations. Strong \textbf{reasoning} abilities (Fig. \ref{fig: intro}(b)) are essential for LLMs to understand these contexts and tools, generating logical thoughts for the next steps. After generating a thought, selecting the appropriate tools from a given list is crucial, demanding effective \textbf{retrieval} skills (Fig. \ref{fig: intro}(c)). Additionally, integrating the correct parameters requires the \textbf{understanding} ability (Fig. \ref{fig: intro}(d)) to interpret tool documentation and corresponding thoughts. Finally, executing the tool-calling action mandates adept \textbf{instruction following} skills (Fig. \ref{fig: intro}(e)) to formulate precise requests for the relevant APIs.
Each tool call executed by LLM must be evaluated to ensure the response meets the intended objective, especially when tools might be unavailable or not perform as anticipated. This crucial evaluation, named the \textbf{review} ability (Fig. \ref{fig: intro}(f)), involves examining tool responses and ascertaining if adequate information has been obtained to resolve the query.

In summary, thorough analyses of each dimension are vital for a comprehensive evaluation of tool-utilization capabilities. Therefore, we introduce \shortname, a framework that decomposes the multi-step tool-calling process into fundamental abilities, evaluating them individually for a more nuanced understanding of tool utilization.

\begin{figure*}[!t]
    \vspace{-4mm}
    \centering
    \includegraphics[width=2.1\columnwidth]{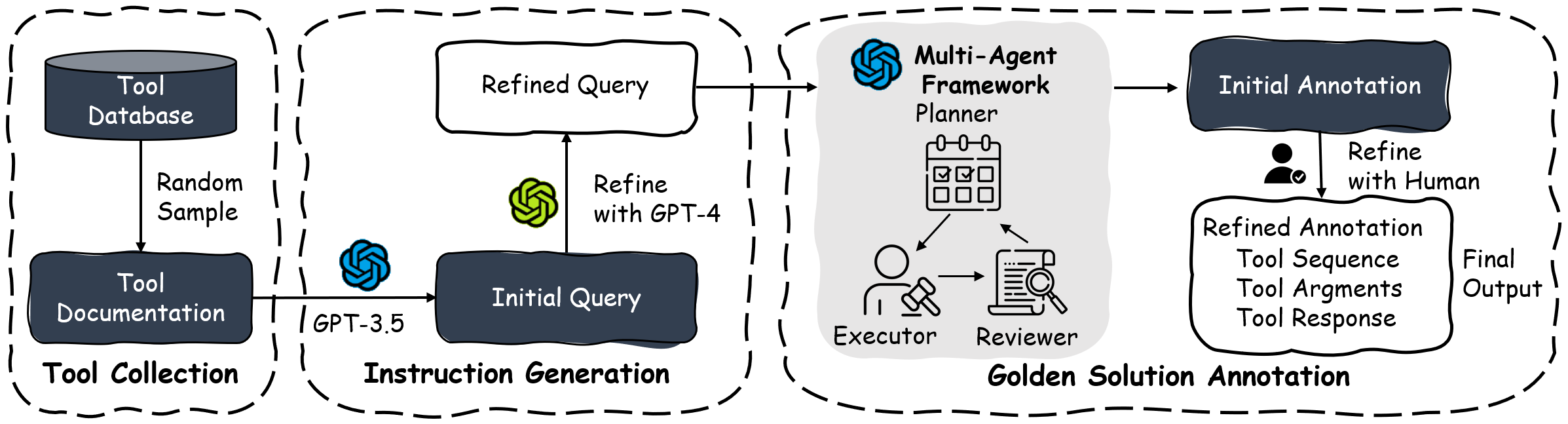}
    \caption{\textbf{Overview of the dataset construction process.} By randomly sampling tools from the tool database, we prompt GPT-3.5 to generate initial queries and further refine them with GPT-4. After that, we develop a multi-agent framework to resolve queries with the provided tools, collecting both solution paths and tool responses. Finally, human experts are employed to verify the annotations and pick high-quality samples.}
    \label{fig:anno_frame}
    \vspace{-2.5mm}
\end{figure*}

\subsection{Fine-Grained Evaluation Protocol}
\label{sec:eval_protocol}

\shortname~takes all the ability dimensions as mentioned above (\textsc{plan, reason, retrieve, understand, instruct}, and \textsc {review}) into consideration, measuring not only the overall performance of tool-utilization but also detailed scores in each dimension.

\subsubsection{Definition}

To formalize, this paper considers a piece of query data as a tuple $(T, q)$, where $T = [tool_1, ..., tool_k]$ is the tool list and $q$ is the query. For each query data piece $(T, q)$, we define the solution path $S=\left[(t_i, a_i, o_i, r_i)\right]_1^{n}$ for the query $q$ as a sequence of thought($t$)-action($a$)-observation($o$)-review($r$) pair along with the final answer $A$, where $t_i, a_i, o_i, r_i$ denotes the thought, the tool-calling action, the observation (\textit{i.e.,} the tool response), and the review on the response at step $i$. Moreover, an action is regarded as a pair $(tool, args)$, where $tool$ is the tool name and $args$ is the parameters to call the tool. Besides the solution path, a plan for a query data piece is defined as a sequence $P = [a_1, ..., a_n]$ donating the proposed action sequence to call at each step.

\subsubsection{Single-Index Evaluation}
\label{sec:single_index_eval}

We create individual metrics to quantitatively analyze LLM abilities for each dimension of tool utilization. Here we describe the measurement of each dimension, and the detailed metric function can be found in Appendix \ref{app: Detailed Evaluation Metrics}.
\begin{itemize}	
    \item \textbf{\textsc{Plan}}: Given a tool list $T$ and query $q$, the LLM is asked to generate a proposed tool-calling action sequence $P^{pred} = [a_1^{pred}, a_2^{pred}, ..., a_{n}^{pred}]$, where $a_i^{pred}$ is the predicted LLM action at step $i$. The planning evaluator then compares $P^{pred}$ with the golden answer $P^{gt}$ by matching actions in both sequences using Sentence-BERT \cite{reimers2019sentence} for similarity calculation and Hopcroft-Karp matching \cite{hopcroft1973n} for maximal similarity pairing. The planning score is measured by the length of the longest-ordered action sequence in the similarity pairing.
    \item \textbf{\textsc{Reason}}: Given a tool list $T$, query $q$, and a prefix of the solution path, the reasoning evaluator asks the LLM to generate the next thought $t_{i+1}^{pred}$. The similarity between $t_{i+1}^{pred}$ and the golden answer $t_{i+1}^{gt}$ is then measured.
    \item \textbf{\textsc{Retrieve}}: Given a tool list $T$, query $q$, and a prefix of the solution path, the retrieval evaluator asks the LLM to choose the next tool $tool^{pred}$ to call, comparing it with the golden answer $tool^{gt}$.
    \item \textbf{\textsc{Understand}}: Given a tool list $T$, query $q$, the understanding evaluator asks the LLM to generate appropriate parameters $args^{pred}$ for the next step and then compares them with the golden answer $args^{gt}$ for similarity.
    \item \textbf{\textsc{Instruct}}: Given a thought $t_i$ with the desired tool and parameters, the LLM is tasked with generating a tool-calling request in a specified format. The evaluator then calculates the accuracy of the tool name and parameter values in this format.
    \item \textbf{\textsc{Review}}: Given a thought $t_i$ and a tool response $o_i$, the LLM is tasked with judging whether the tool response successfully achieves the goal mentioned in the thoughts. It must also determine the type of errors if the goal is not achieved. This evaluation is conducted as a multiple-choice problem with five options: \textit{Success}, \textit{Internal Error}, \textit{Input Error}, \textit{Irrelevant Response}, and \textit{Unable to Accomplish}.
\end{itemize}

\subsubsection{End-to-End Evaluation}
\label{sec:overall_eval}
End-to-end evaluation requires LLMs to generate the whole solution path $S^{pred}$ as well as the final answer $A^{pred}$ given one query data piece. We adopt the win rate proposed in ToolBench~\cite{qin2023toolllm} to gauge the overall performance, which evaluates LLM abilities by comparing their response quality against that of GPT-3.5. The result shows that our single-index evaluation is consistent with the overall performance.

\subsection{Dataset Construction}
\label{sec:data_construct}

The construction of \shortname~consists of three main phases: \textit{tool collection}, \textit{instruction generation}, and \textit{golden solution annotation}. The overview of the construction is shown in Fig. \ref{fig:anno_frame}.

\subsubsection{Tool Collection}

The collection quality of tools has a direct impact on instruction generation and tool utilization evaluation. We follow two principles during the collection process:\\
\textbf{$\bullet$ High Availability and Usage Rate}. Considering that \shortname~is expected to cover most daily and practical use cases, we carefully select $1\sim 2$ tools for each specific domain, including Research, Travel, Entertainment, Web, Life, and Financials, resulting in $15$ tools as our basic tool set. \\ 
\textbf{$\bullet$ Complete Documentation}. Despite the numerous tools collected in ToolBench \cite{qin2023toolllm} from RapidAPI, the documentation quality is not guaranteed. To reduce the failure of tool-calling cases caused by inadequate tool descriptions, which focus the evaluation attention on pure LLM abilities, we manually generate high-quality and detailed tool documentation for each tool.

\subsubsection{Instruction Generation}

The testing instructions determine the practicality and difficulty of our evaluation. To guarantee the diversity of the queries, we uniformly sample $2\sim 3$ tools each time and prompt GPT-3.5 to generate $N$ instructions $Q=\{q_1, ..., q_N\}$ that need these tools. Concretely, the prompt consists of three parts: (1) the instruction that requests LLM to generate corresponding queries, (2) detailed tool documentation, and (3) few-shot examples. We randomly shuffle the tool documentation list and select different few-shot examples each time, so that the LLM can pay different attention to the text thereby encouraging the model to create wide-ranging instructions. After that, the stronger GPT-4 is utilized to revise and refine the generated instructions, aiming to further enhance the feasibility and diversity. The detailed prompts are listed in the Appendix \ref{app: prompt detail}.

\subsubsection{Golden Solution Annotation}

Annotating the solution path manually to various queries is labor-intensive and unable to scale up the dataset quickly. To overcome this problem, we leverage a novel multi-agent paradigm plugging with simple human verification to resolve the complicated and massive solution annotations. Specifically, instead of instantiating only one LLM to handle the whole solution annotation path, we explicitly disentangle the annotation task into three different functionalities, including planner, executor, and reviewer: the planner decides what should be done in the next step; the executor is responsible for generating the exact tool name as well as its parameters and executing the tool to obtain the response. The reviewer is assigned to revise the response from the tool and judge if the task is finished given the external feedback. 

Thanks to the decomposition of functionalities, each agent can accomplish its duty without switching the role required by each step, therefore, significantly reducing the error generation during the annotation process compared to conventional CoT \cite{wei2022chain} or ReAct \cite{yao2022react} approaches. To guarantee the quality of the calling process, we adopt GPT-3.5 to accomplish the whole machine annotation phase. In the end, human annotators are employed to review the solution chain and manually filter invalid instruction-solution pairs.



\subsubsection{Inclusive Difficulty Evaluation}

During the evaluation, we empirically found that a few LLMs, especially small-scale ones, demonstrate poor ability in instruction following, therefore yielding responses that are unparseable with the pre-agreed format in the instruction prompt. Due to the large amount of parse failures on the response, the evaluation score can get distorted, losing the authenticity to reflect the real ability of the model.

To address the issue, we carefully designed the instruction task prompt and the evaluation granularity with both easy and difficult levels, providing inclusive evaluation on most language models. Specifically, the easy level adheres to a simple string format and focuses more on the semantic quality of the text, while the difficult level adopts JSON format, which is more commonly used in products (\eg, the JSON mode\footnote{\href{https://openai.com/blog/new-models-and-developer-products-announced-at-devday}{https://openai.com/blog/new-models-and-developer-products-announced-at-devday}} in GPT-4~\cite{openai2023gpt4}), and conducts a more strict, fine-grained evaluation of the response content, \eg, exact match on tool name and parameters.

\subsection{Dataset Summary}

To this end, we generate 1,500 initial instruction-solution pairs and pick 553 after two-round human verifications. We extract the desired information required by each evaluation protocol to construct the respective \textsc{Instruct, Retrieve, Plan, Reason, Understand, Review} subsets for \shortname~benchmark, resulting in 23,305 test cases in total (Refer to Appendix \ref{app: Dataset Statistics} for more detailed statistics of \shortname).




\label{sec:data_summary}

%% file: sections/3_experiments.tex
\section{Experiments}


\input{tables/main_table.tex}

\subsection{Experimental Setup}
\label{sec:exp_setup}

We evaluate both API-based commercial and open-source LLMs on \shortname, with a total number of $20$ models, aiming to provide a comprehensive benchmark for current large language models. \\
(1) For API-based LLMs, we select three representative models: GPT-3.5 and GPT-4 from OpenAI, and Claude2 from Anthropic. \footnote{Our experiments are conducted between 12/01/2023 and 12/10/2023. The version for GPT-4 is \texttt{gpt-4-1106-preview}, for GPT-3.5 is \texttt{gpt-3.5-turbo-16k}, and for Claude2 is \texttt{claude-2.1}.}\\
(2) For open-source LLMs, we choose a wide spectrum of models, including LLaMA2 \cite{touvron2023llama2}, CodeLLaMA \cite{roziere2023code}, QWen \cite{bai2023qwen}, InternLM \cite{team2023internlm}, Baichuan2 \cite{yang2023baichuan},  WizardLM \cite{xu2023wizardlm}, Vicuna \cite{chiang2023vicuna}, AgentLM \cite{zeng2023agenttuning}, Mistral \cite{jiang2023mistral} and ChatGLM3 \cite{zeng2022glm}.

\subsection{Main Results}
\label{sec:main_result}

The detailed experimental results are shown in Tab. \ref{tab:main_results}. 
In this section, We aim to answer three research questions below.

\begin{figure}[!t]
    \includegraphics[width=1.0\columnwidth]{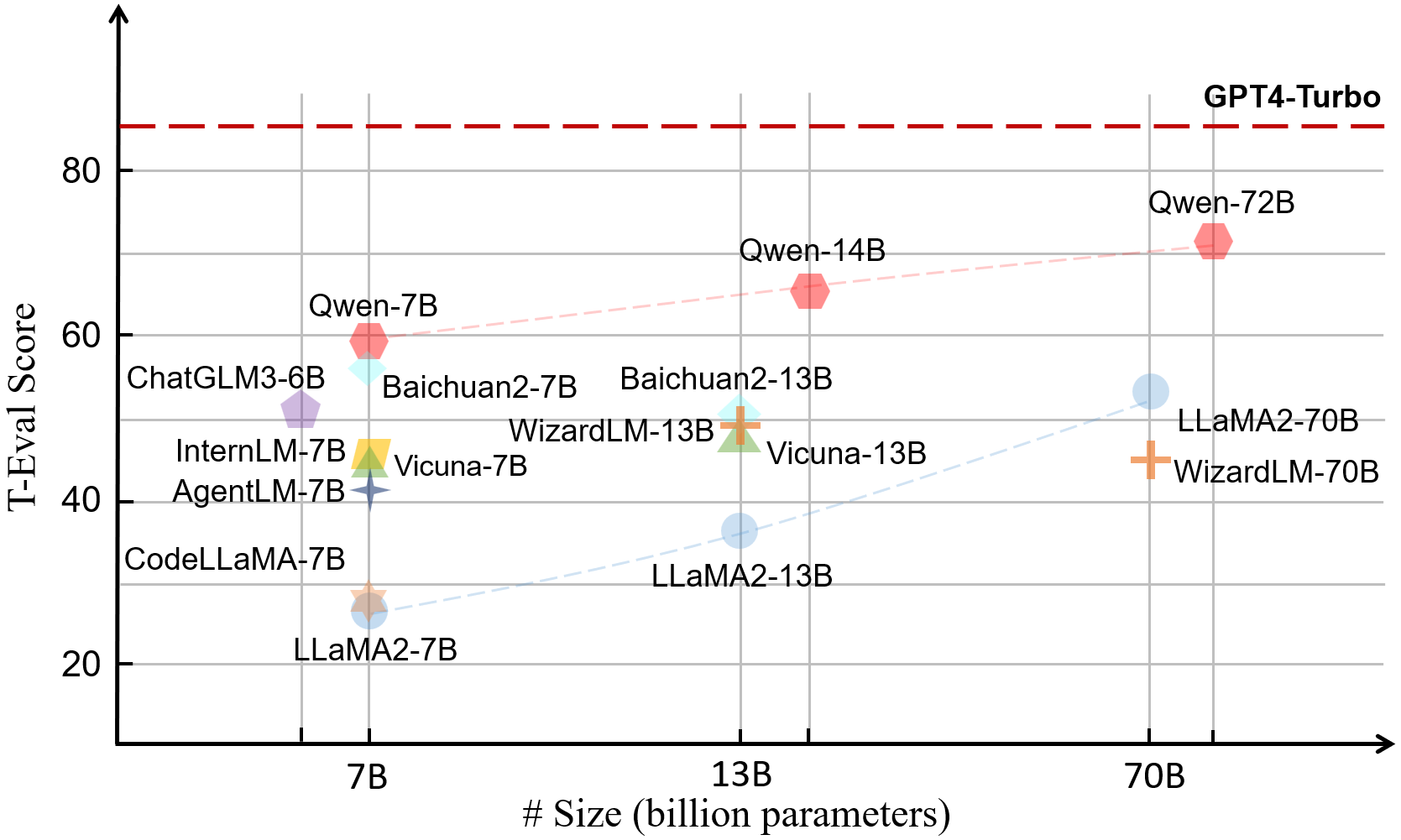}
    \caption{\textbf{\shortname~scale v.s the size of models}. Both LLaMA2 and Qwen strengthen their tool utilization abilities as the models scale up. However, there still exists a clear performance gap between open-source models and GPT-4.}
    \label{fig:scale}
\end{figure}

\noindent\textbf{Q1: Which Model is Better at Tool Utilization?} The results in Tab. \ref{tab:main_results} show that GPT-4 achieves the highest score, with an overall score of 86.4, setting the pilot of the well-instructed and skillful tool-utilization LLMs. Apart from GPT-4, API-based commercial LLMs, including GPT-3.5 and Claude2, get competitive scores on both string and JSON formats, indicating their strong abilities in acting as tool agents.

As for open-source models, we evaluate models with three different scales: around 7B, 13B, and 70B. It can be concluded from Fig. \ref{fig:scale} that the performance of the model monotonically increases as the model scale increases.
Among them, Qwen-7B gets the best of two worlds in terms of the model sizes and evaluation scores. With 7 billion parameters, Qwen-7B exhibits a competitive ability to understand complicated instructions and reply in a strict format in JSON. When scaled to 72B, the overall score of Qwen rises to 71.4\%, significantly reducing the gap between open-source and API-based models. We attribute this to the training on a human-in-loop self-instruction dataset, which encompasses high-quality format-specific instructions generated by Qwen team \cite{bai2023qwen}.

\noindent\textbf{Q2: How Far Are We from Skillful Tool Agents?} By explicitly disentangling the evaluation through model abilities, we can gain a deeper understanding of the pros and cons of current LLMs, providing new perspectives in developing better tool agents.

First, open-source LLMs lack the instruction-following ability to respond with specific formats, which is the very first step to constructing well-regulated and high-usable tool agents. Without a legal response format, the system can not successfully extract the information generated by LLMs by a constant protocol, not to mention the correct tool executions. However, only a small amount of models achieve both high scores on \textsc{Instruct} subset under string and JSON format. Besides, there are still large performance gaps between string and JSON evaluation protocols on other subsets. Considering that the \textit{understand} ability of the Qwen-72B is comparable with GPT-4 evaluated in string format (84.5 vs 83.2), its JSON format result is more than 20 points lower, pinpointing the necessity to enhance the ability of open-source LLMs solving problems with specific formats.

Second, tool retrieval presents a relatively challenging task for most LLMs. Even the largest open-source model, Qwen-72B only reaches 65.0\%, which is more than 20 points lower than GPT-3.5.

Lastly, compared to planning, LLMs are more likely to perform worse in reviewing the status of API responses (\textit{i.e.,} review in \shortname), which is a core capability to interact with the dynamic environment when acting as an agent. Most models only reach 50\%$\sim$60\%, compared to 95\% achieved by GPT-4, indicating that more attention should be paid to the review ability of current LLMs.

\noindent\textbf{Q3: What Makes for Good Training Data for Tool Utilization?}
Supervised finetuning is an efficient and necessary practice in empowering LLMs with certain abilities on downstream tasks. Recent research \cite{ouyang2022training, touvron2023llama2} finds that high-quality dialogue-style instruction data is the key to well-instructed LLMs. However, the analysis of what makes for good training data for tool utilization is under-explored. We identify two types of training data: \textit{(1) general instruction following data}, and \textit{(2) task-specific tuning data}. 

As for high-quality general instructions, Vicuna adopts user-shared conversations collected from ShareGPT\footnote{\href{https://sharegpt.com/}{https://sharegpt.com/}}, and WizardLM uses complex and diverse instructions by evolving existing data in depth and width. Both of them are trained starting from LLaMA2, providing natural ablation on the effectiveness of high-quality (diverse and complex) instructions to the tool learning ability of LLM. When the model scale is small (7B), these data types enhance the model considerably. However, the increments diminish (even worse) as the model scales up (see WizardLM-70B). This further indicates that simply scaling the model scale does not always bring improvements, proper training data also matter in the scaling law.

In terms of task-specific tuning data, we select two typical types of corpus: code and agent, corresponding to CodeLLaMA and AgentLM, respectively. Compared to CodeLLaMA, which uses code data, AgentLM obtains better scores, showcasing that agent-related data may bring more benefits to tool learning. However, neither CodeLLaMA nor AgentLM appears to have significant advantages to Vicuna, indicating the necessity of high-quality instruction following data for tool utilization.

\section{Discussion}

\subsection{Format Following v.s Problem Solving}
Format following is an essential ability of LLMs, \textit{i.e.,} reply with a specific format when executing certain tasks. There are massive efforts devoted to enhancing this ability when using LLM as agents \cite{zhou2023instruction,xu2023wizardlm}. In our experiments, we find that this ability may need to be acquired jointly with problem-solving.
For instance, both ChatGLM3-6B and Baichuan2-7B obtain roughly 80\% with the JSON evaluation protocol on \textsc{Instruct} subset, which reveals them holding a strong ability in JSON format rendering. However, they struggle to generate valid JSON format responses on \textsc{Plan} and \textsc{Reason} subsets, \textit{i.e.,} the divergences of JSON and string scores are quite large on these subsets. Such a phenomenon suggests that the ability to output specific formats, \textit{i.e.,} JSON, does not guarantee the ability to resolve all the problems with this format. This further indicates that one should integrate the requested format into the tasks and train them jointly so that the model can understand and behave well under certain protocols.

\subsection{Inclusive Evaluation Protocol}

From Tab. \ref{tab:main_results}, we can observe that quite a few amount of open-source LLMs struggle on the JSON evaluated protocols, especially on the \textsc{Reason}, \textsc{Retrieve} and \textsc{Understand} subsets. Although the JSON format evaluation best approximates the real use cases of tool agents, it fails to provide hierarchy discriminations across various models when they are not adept at specific instructions requested by the task. For instance, Baichuan2-13B exhibits poor abilities in JSON format instruction, which leads to low scores under the JSON evaluations in \textsc{Plan} subset. However, it has little relationship with the basic ability of model planning, since it achieves 65.6\% (25 points larger) when evaluated with string format.
This validates the necessity to provide continual difficulty level evaluation protocols \cite{schaeffer2023emergent}, otherwise, one can get little understanding of the detailed abilities of LLMs on this benchmark but a low score simply due to the incorrect format, especially for weak models. In \shortname, by seamlessly converting the strict format matching into semantic sentence analysis, our inclusive evaluation protocol gets rid of the inflexible measurement, unearthing the inner capability of the model. 



\subsection{Comparison to Other Benchmarks}
\label{sec:peval_validate}
\begin{figure}[!t]
    \includegraphics[width=1.0\columnwidth]{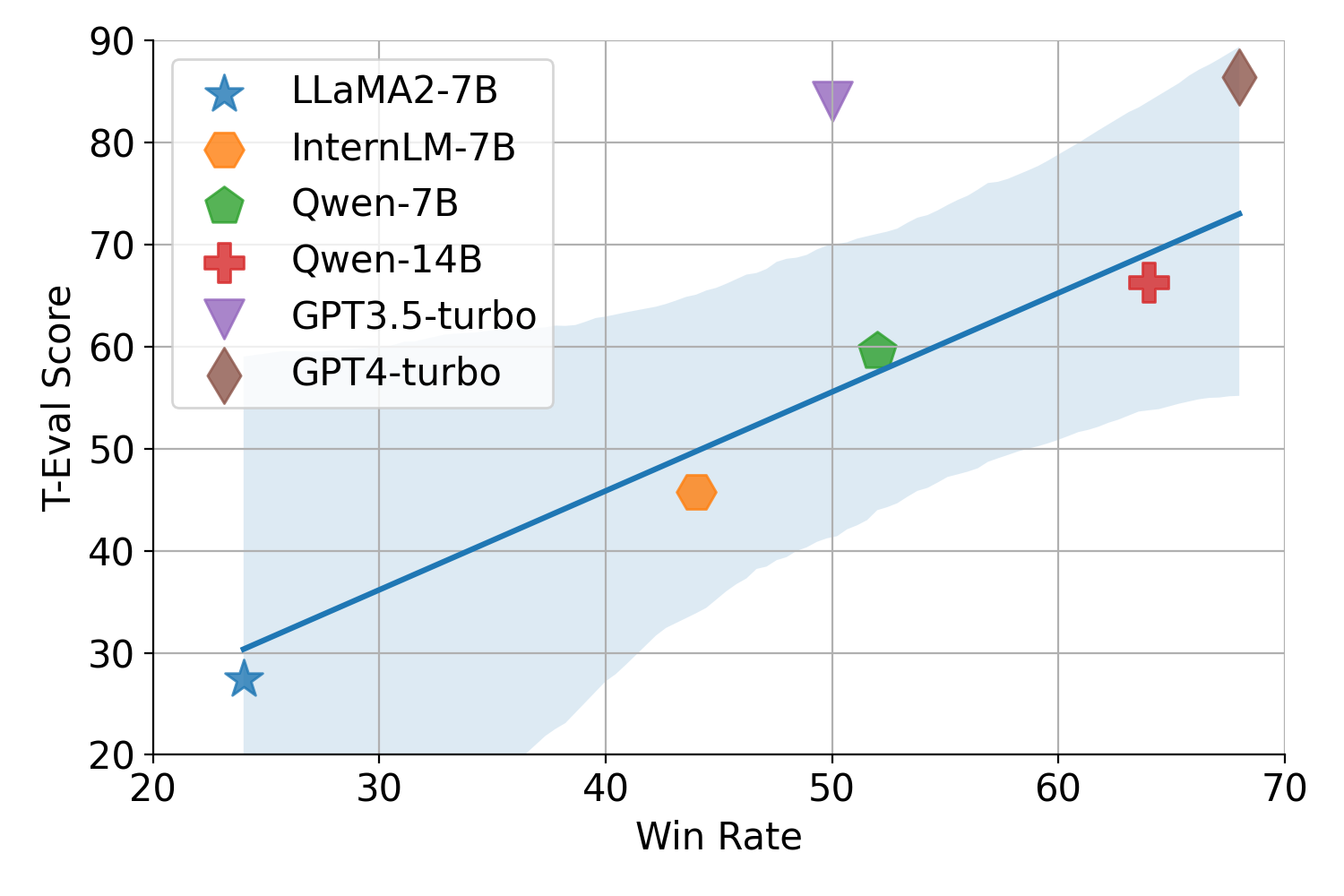}
    \caption{\shortname~average score v.s Win Rate proposed in ToolBench \cite{qin2023toolllm} on several representative LLMs. \shortname~score (objective) demonstrates similar trends with Win Rate (judged by GPT-4).}
    \label{fig:winrate}
\end{figure}

We compare our fine-grained evaluation protocols with existing tool evaluation approaches and investigate if they show the same trends as ours. We adopt the win rate proposed in ToolBench \cite{qin2023toolllm} as the representative holistic evaluation method, and evaluate several open-source models by comparing the response quality with GPT-3.5-turbo, judged by GPT-4. The results are shown in Fig. \ref{fig:winrate}. We can find that the holistic evaluation reveals similar trends in these models with that in \shortname, which validates the reasonability and generalization of our benchmark. When taking a close look at the results, we can observe that Qwen-7B achieves a 52\% win rate over GPT-3.5-turbo response. However, there still exists a gap between Qwen-7B and GPT-3.5 in various ability domains under human judgment \cite{zheng2023judging}, which implies that holistic evaluation is sometimes inaccurate. Our evaluation protocol not only reflects such divergence clearly, but also showcases the details of abilities in tool utilization, suggesting that \shortname~to be a more rational and comprehensive evaluation benchmark for tool utilization. 

%% file: tables/main_table.tex
\newcommand\redcolor[1]{\cellcolor{gray!30!red!30}{#1}}
\newcommand\bluecolor[1]{\cellcolor{gray!30!blue!30}{#1}}
\begin{table*}[ht]
    \caption{\textbf{Main Results of \shortname.} Overall stands for the score calculated from an average of metrics on all subsets. (\textbf{bold} denotes the best score among all models, and \underline{underline} denotes the best score under the same model scale.)}
    \label{tab:main_results}
    \setlength{\tabcolsep}{4pt}
    \renewcommand{\arraystretch}{1.1}
    \centering
    \resizebox{\textwidth}{!}{%
    \begin{tabular}{l|ccccccccccc|c}
    \toprule
    \multirow{2}{*}{\textbf{Model}} & \multicolumn{2}{c}{\textbf{\textsc{Instruct}}} & \multicolumn{2}{c}{\textbf{\textsc{Plan}}} & \multicolumn{2}{c}{\textbf{\textsc{Reason}}} & \multicolumn{2}{c}{\textbf{\textsc{Retrieve}}} & \multicolumn{2}{c}{\textbf{\textsc{Understand}}} & {\textbf{\textsc{Review}}} &  \multirow{2}{*}{\textbf{Overall}}\\
    \addlinespace[1pt]
    & \small{String} & \small{JSON} & \small{String} & \small{JSON} & \small{String} & \small{JSON} & \small{String} & \small{JSON} & \small{String} & \small{JSON} & \small{Choice} & \\ %
    \midrule
\multicolumn{13}{l}{\redcolor{\textbf{\textit{API-Based}}}} \\
Claude2 & \textbf{97.7} & 97.8 & 87.1 & 84.9 & 62.9 & 62.8 & 76.5 & 78.2 & 74.9 & 82.0 & 70.4 & 78.8 \\
GPT-3.5 & 94.1 & \textbf{99.1} & 86.6 & 86.6 & 65.2 & \textbf{70.3} & \textbf{98.3} & 86.2 & 82.9 & 88.1 & 75.6 & 84.0 \\
GPT-4 & 96.7 & 95.9 & \textbf{88.9} & \textbf{86.7} & \textbf{65.6} & 65.1 & 91.3 & \textbf{86.6} & \textbf{83.2} & \textbf{88.3} & \textbf{94.5} & \textbf{86.4} \\
\midrule
\multicolumn{13}{l}{\bluecolor{\textbf{\textit{Open-Sourced}}}} \\
LLaMA2-7B & 68.7 & 0.2 & 47.0 & 9.1 & 37.1 & 7.1 & 30.3 & 3.5 & 36.8 & 12.1 & 38.6 & 27.4 \\
CodeLLaMA-7B & \underline{\textbf{96.0}} & 0.9 & 61.4 & 44.3 & 28.7 & 0.9 & 3.6 & 1.2 & 25.4 & 1.4 & 40.0 & 28.6 \\
AgentLM-7B & 80.8 & 13.1 & 53.1 & 15.9 & 50.1 & 17.5 & 70.2 & 13.8 & 66.4 & 26.1 & 44.8 & 41.4 \\
Vicuna-7B & 65.3 & 30.8 & 13.4 & 47.8 & 47.6 & \underline{49.9} & 12.4 & 32.6 & 66.8 & \underline{54.2} & 58.5 & 44.8 \\
InternLM-7B & 48.4 & 29.9 & 67.7 & 43.1 & 48.8 & 25.0 & 72.1 & 22.2 & 70.4 & 30.2 & 46.2 & 45.8 \\
ChatGLM3-6B & 63.3 & 80.8 & 46.9 & 38.5 & 48.2 & 24.1 & 66.5 & 24.0 & 79.9 & 35.6 & 54.8 & 51.4 \\
Mistral-7B & 59.7 & 63.6 & \underline{77.2} & \underline{64.9} & \underline{63.0} & 15.3 & \underline{92.6} & 11.0 & 79.8 & 18.1 & \underline{63.2} & 56.0 \\
Baichuan2-7B & 68.0 & 78.0 & 65.6 & 39.0 & 51.3 & 31.3 & 73.7 & 28.5 & \underline{80.1} & 39.2 & 61.4 & 56.5 \\
Qwen-7B & 28.7 & \underline{94.2} & 66.2 & 63.1 & 56.4 & 34.1 & 89.0 & \underline{35.3} & 77.7 & 46.1 & 61.6 & \underline{59.5} \\
\midrule
LLaMA2-13B & 66.7 & 0.0 & 48.7 & 65.1 & 42.4 & 10.5 & 42.7 & 6.6 & 45.7 & 13.1 & 53.0 & 37.3 \\
Vicuna-13B & \underline{67.0} & 30.8 & 25.8 & 54.0 & 56.3 & \underline{49.1} & 19.8 & 20.9 & 73.0 & 58.8 & 60.8 & 48.1 \\
WizardLM-13B & 14.1 & 65.5 & 77.7 & 40.8 & 36.0 & 25.2 & 68.5 & 22.2 & 64.0 & 31.6 & \underline{71.5} & 49.0 \\
Baichuan2-13B & 8.0 & 51.7 & 69.5 & 52.1 & 56.6 & 27.1 & 84.5 & 26.9 & \underline{80.5} & 31.5 & 57.3 & 50.3 \\
Qwen-14B & 49.7 & \underline{97.6} & \underline{79.6} & \underline{69.7} & \underline{58.6} & 46.1 & \underline{\textbf{95.9}} & \underline{55.3} & 65.0 & \underline{\textbf{64.3}} & 56.9 & \underline{66.3} \\
\midrule
WizardLM-70B & 9.6 & 31.7 & 81.5 & 42.7 & 38.4 & 47.0 & 38.3 & 56.2 & 66.1 & 61.1 & 28.7 & 44.2 \\
LLaMA2-70B & \underline{84.5} & 73.4 & 58.0 & 63.1 & 44.7 & 17.5 & 62.0 & 17.1 & 67.3 & 22.3 & 62.8 & 53.0 \\
Qwen-72B & 27.8 & \underline{\textbf{98.3}} & \underline{\textbf{85.1}} & \underline{\textbf{73.4}} & \underline{\textbf{63.5}} & \underline{\textbf{55.4}} & \underline{76.8} & \underline{\textbf{65.0}} & \underline{\textbf{84.5}} & \underline{66.1} & \underline{\textbf{80.3}} & \underline{\textbf{71.4}} \\
    \bottomrule
\end{tabular}}
\end{table*}

%% file: sections/4_related_work.tex
\section{Related Work}

\paragraph{Augmenting LLMs with Tools} There are two paradigms to empower LLMs with external tools, and the first one is regarding external tools as specific tokens and fine-tuning parts or full of the model \cite{schick2023toolformer, parisi2022talm, lewis2020retrieval, hao2023toolkengpt}. However, these methods need a large amount of tool-relevant data and struggle to adapt newly appeared tools. Recently, the strong in-context learning ability \cite{brown2020language} promotes researchers to focus more on the second paradigm, which is augmenting LLMs with tools by giving in-context tool descriptions and demonstrations \cite{hsieh2023tool, mialon2023augmented, ruan2023tptu, patil2023gorilla}. This paradigm has achieved great tool-calling potentiality and resulted in successful applications such as ChatGPT plugins. \shortname~focus on this paradigm and evaluate scores of various foundation models.

\paragraph{Evaluating LLMs}
LLM evaluation is essential to ensure that LLM can be effective in understanding and generating human-preferred text and reliable for deployment in real-world applications \cite{guo2023evaluating, chang2023survey}. Many benchmarks have been established to evaluate base abilities on question-answering tasks \cite{rajpurkar2016squad, clark2018think, glockner2018breaking}, natural language understanding tasks \cite{wang2018glue, wang2019superglue, hendrycks2020measuring}, and commonsense reasoning tasks \cite{lu2022learn}. Recently, LLM evaluation has extended towards specific directions like code generation \cite{chen2021evaluating, austin2021program, du2023classeval} and hallucination \cite{li2023halueval, chen2023benchmarking}. Some benchmarks also test the performance of LLM-based agents in a wide range of scenarios \cite{liu2023agentbench, wang2022scienceworld}.

Several benchmarks exist for evaluating tool utilization, focusing primarily on aspects of response comparison (e.g., ToolQA \cite{zhuang2023toolqa}), tool call accuracy (e.g., Gorilla \cite{patil2023gorilla}), or a combination of both (e.g., API-Bank \cite{li2023api}). ToolBench \cite{qin2023toolllm} introduces a novel approach by employing an LLM as a judger to assess the overall solution path. Furthermore, the study in \cite{qin2023tool} investigates the performance improvement attributable to tool utilization. Different from above, \shortname~emerges as the first benchmark dedicated to the fine-grained evaluation of tool utilization capabilities.

\paragraph{Prompting LLMs as Agents} Prompting enhances the reasoning capabilities of LLMs by providing instructions or examples. Techniques such as the Chain of Thought (CoT) and Tree of Thought (ToT) \cite{wei2022chain, yao2023tree} encourage LLMs to engage in comprehensive thinking for more accurate reasoning. Advanced systems like ReAct, ReWOO, SwiftSage, DyLAN, and DP-LLM \cite{yao2022react, xu2023rewoo, lin2023swiftsage, liu2023dynamic, dagan2023dynamic}  further develop LLM agents. These systems use advanced prompting methods to guide LLMs, unleashing the potential of models. 

%% file: sections/5_conclusion.tex
\section{Conclusion}

In this paper, we propose \shortname, a comprehensive and fine-grained tool utilization evaluation benchmark for LLMs. \shortname~explicitly disentangles the tool utilization tasks along the model ability, with dedicated evaluation protocols designed for respective tasks, unearthing the real ability of the evaluated models. Such a step-wise evaluation delivers a thorough analysis and pinpoints the main bottlenecks of current LLMs in tool learning, providing valuable insights into further development of tool agents. 

%% file: sections/99_appendix.tex
\definecolor{lightgray}{gray}{0.95}
\lstdefinestyle{prompt}{
    basicstyle=\ttfamily\fontsize{7pt}{8pt}\selectfont,
    frame=none,
    breaklines=true,
    backgroundcolor=\color{lightgray},
    breakatwhitespace=true,
    breakindent=0pt,
    escapeinside={(*@}{@*)},
    numbers=none,
    numbersep=5pt,
    xleftmargin=5pt,
}
\tcbset{
  aibox/.style={
    top=10pt,
    colback=white,
    colframe=black,
    colbacktitle=black,
    enhanced,
    center,
    attach boxed title to top left={yshift=-0.1in,xshift=0.15in},
    boxed title style={boxrule=0pt,colframe=white,},
  }
}

\section{\shortname~Benchmark Details}

\subsection{Dataset Statistics}
\label{app: Dataset Statistics}
\begin{figure}[!ht]
    \includegraphics[width=1.0\columnwidth]{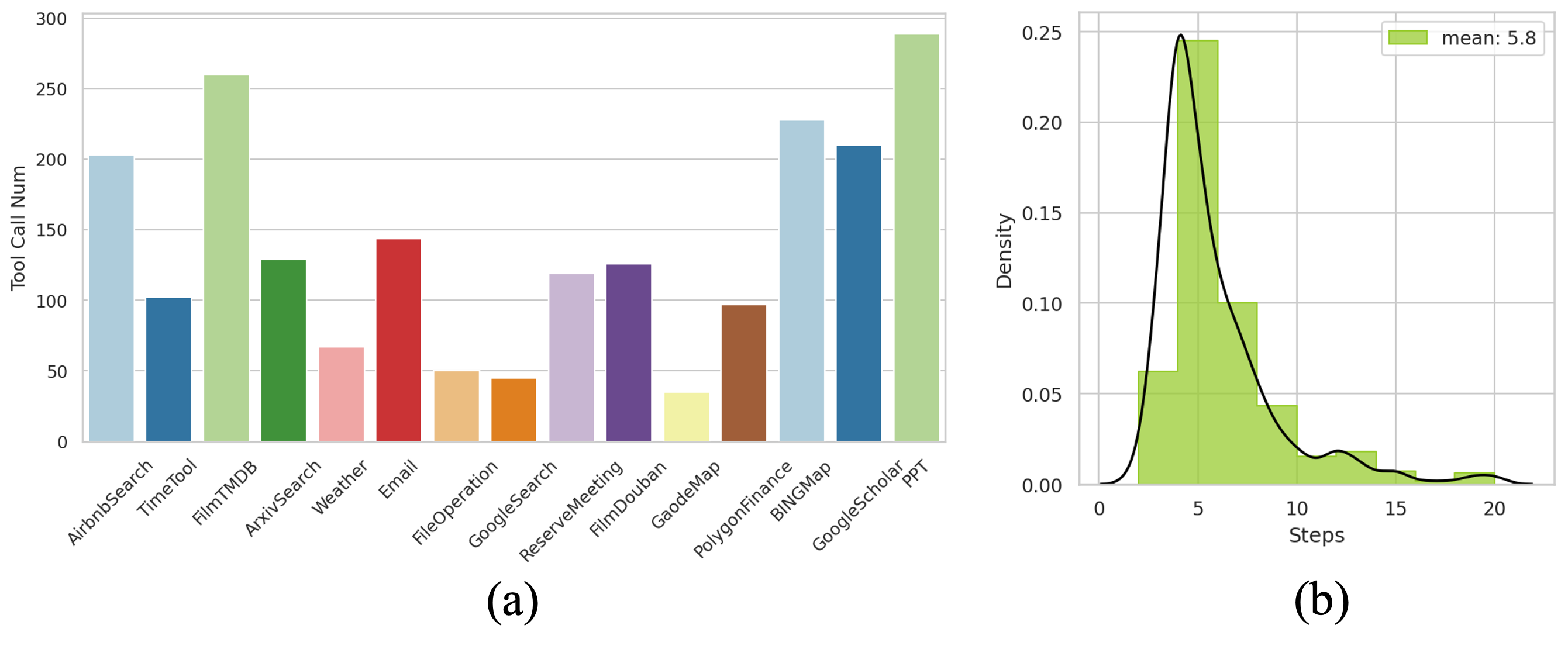}
    \caption{(a) Tool calling categorical distribution and (b) tool calling step distribution accumulated in the whole annotation paths in \shortname.}
    \label{fig:step_dist}
\end{figure}

\shortname~ originates from 533 high-quality query-solution annotation pairs, consisting of 23,305 test cases in total, ranging from \textsc{Instruct, Plan, Reason, Retrieve, Understand} and \textsc{Review} subsets. Detailed statistics of each subset are shown in Tab. \ref{tab:data_stat}. We also visualize the distribution of tool calling steps in the whole annotation paths in Fig. \ref{fig:step_dist}. \shortname~covers all tool sets and yields 5.8 average calling steps for each query, validating the generalization and discrimination for tool utilization evaluation.

\input{tables/new_dataset.tex}

\section{Implementation Details}

\textbf{Experimental Details.} 
To evaluate the pure ability of the single model, we adopt ReAct \cite{yao2022react} as the basic agent paradigm for end-to-end evaluation and limit the maximum action step to 20 to ensure the efficient and accurate question-solving ability of LLM. As for the single-index evaluation, we prompt with a multi-turn conversation style to the LLM and gauge the response. If not specified, we choose the `chat'/`instruct' version of open-sourced models for evaluation.


\label{app: prompt detail}
\subsection{Prompts Demonstration}

Please refer to the respective prompt block for a detailed demonstration.

\subsubsection{Query Generation}
The corresponding prompt is presented in Fig. \ref{fig: Query Generation}.

\subsubsection{Query Refinement}
The corresponding prompt is presented in Fig. \ref{fig: query refinement}.

\subsubsection{Multi-Agent Annotation Prompt}
The corresponding prompt is presented in Fig. \ref{fig: multi-agent annotation prompt}.

\subsection{Dataset Demonstration}

Please refer to the respective prompt block for each detailed dataset demonstration.

\subsubsection{\textsc{Instruct}}
The corresponding prompt is presented in Fig. \ref{fig: dataset demonstration instruct}.

\subsubsection{\textsc{Plan}}
The corresponding prompt is presented in Fig. \ref{fig: dataset demonstration plan}.

\subsubsection{\textsc{Reason}}
The corresponding prompt is presented in Fig. \ref{fig: dataset demonstration reason}.

\subsubsection{\textsc{Retrieve}}
The corresponding prompt is presented in Fig. \ref{fig: dataset demonstration retrieve}.

\subsubsection{\textsc{Understand}}
The corresponding prompt is presented in Fig. \ref{fig: dataset demonstration understand}.

\subsubsection{\textsc{Review}}
The corresponding prompt is presented in Fig. \ref{fig: dataset demonstration review}.

\section{Detailed Evaluation Metrics}
\label{app: Detailed Evaluation Metrics}
\shortname~ decomposes tool utilization capability into six ability dimensions: \textsc{Instruct, Plan, Reason, Retrieve, Understand} and \textsc{Review}, we carefully designed evaluators and metrics for all dimensions with two difficulty formats: JSON and string. The JSON format asks the LLM to generate standard JSON format responses, while the string format allows the LLM to answer in a relatively loose format. 

Let us recap the formalization of each tool-calling component first: A piece of query data is considered as a tuple $(T, q)$, where $T = [tool_1, ..., tool_k]$ is the tool list with $k$ tools and $q$ is the query. For each query data piece $(T, q)$, the solution path $S=\left[(t_i, a_i, o_i, r_i)\right]_1^{n}$ is defined as a sequence of thought($t$)-action($a$)-observation($o$)-review($r$) pair along with the final answer $A$, where $t_i, a_i, o_i, r_i$ denotes the thought, the tool-calling action, the observation (i.e. the tool response), and the review on the response at step $i$, respectively. Moreover, an action $a$ is regarded as a pair $(tool, args)$, where $tool$ is the tool name and $args$ is the parameters to call the tool. Besides the solution path, a plan for a query data piece is defined as a sequence $P = [a_1, ..., a_n]$ donating the proposed actions to call at each step.

\subsection{\textsc{Instruct}}
The LLM is required to generate a tool-calling request using a specified template, based on the provided tool name and parameters. This request must adhere to a predetermined format, either in JSON or string, with varying template structures. Initially, the evaluator determines if the request meets the format requirements. A \textit{passing score} of $0.5$ is awarded for successfully meeting these format standards. Once the format check is passed, the request is further evaluated for parameter accuracy, with a \textit{parameter score} assigned. This score is calculated as $0.5$ multiplied by the percentage of correctly matched parameters. The final score is the sum of the \textit{passing score} and the \textit{parameter score}.

\subsection{\textsc{Plan}}
The LLM is tasked with generating a plan using a provided list of tools to solve a query. To evaluate the similarity between the predicted plan from the LLM $P^{pred} = [a_1^{pred}, a_2^{pred}, ..., a_{n^{pred}}^{pred}]$ and the gold answer $P^{gt} = [a_1^{gt}, a_2^{gt}, ..., a_{n^{gt}}^{gt}]$ from human annotators, the planning evaluator begins by computing a similarity matrix $S$. This matrix represents the similarity scores for all action pairs $(a_i = (tool_i, args_i), a_j = (tool_j, args_j))$ between the prediction and the golden answer:
$$S_{i,j} = \beta \sigma(tool_i, tool_j) + (1 - \beta) \sigma(args_i, args_j).$$
In this approach, $\sigma$ is the similarity function between two sentences. We employ Sentence-BERT \cite{reimers2019sentence} as $\sigma$, which involves embedding the two sentences and then calculating the cosine similarity between these embeddings as the similarity score. The underlying BERT model used is \texttt{all-mpnet-base-v2}.\footnote{\href{https://www.sbert.net/docs/pretrained_models.html}{https://www.sbert.net/docs/pretrained\_models.html}} Furthermore, $\beta$ is a hyperparameter that determines the relative importance of the tool name in comparison to the tool parameters in similarity calculation. In our implementation, we set $\beta = 0.75$. 

After getting the similarity matrix $S$, a bipartite graph is built where one part is the set of predicted actions and another part is the set of golden answer actions. Two actions are linked if their similarity is greater than a predefined threshold, set at $0.7$ in our implementation. We then employ the Hopcroft-Karp matching algorithm \cite{hopcroft1973n} to compute a max-weighted match from this graph. Subsequently, the Longest Increasing Subsequence (LIS) algorithm is used to determine the longest-ordered action sequence within this max-weighted match. Denoting the length of this sequence as $l$, we calculate the precision and recall as  $p = l / n^{pred}$ and $r = l / n^{gt}$, respectively. The plan score is thus defined as:
$$
\text{plan score} = \frac{2pr}{p + r}.
$$
Regarding the input formats, in the JSON format, the LLM is tasked with generating a list of actions, with each action represented as a dictionary comprising the tool name and its corresponding parameters. Conversely, in the string format, the LLM articulates each action in a separate line.

\subsection{\textsc{Reason}}
\label{app: Reason Detail}
Given a tool list $T$, query $q$, and a prefix of the solution path, the LLM is asked to generate the next thought $t_{i+1}^{pred}$. The similarity between $t_{i+1}^{pred}$ and the golden answer $t_{i+1}^{gt}$ is then measured, and the similarity is calculated by Sentence-BERT (the same as the planning evaluator). In the JSON format, LLMs need to generate a dictionary containing the next thought, as well as the next tool name and the corresponding parameters, which means we evaluate \textsc{Reason}, \textsc{Retrieve}, \textsc{Understand} using the same LLM output. Regarding the string format, LLMs only need to generate the next thought in a single line.

\subsection{\textsc{Retrieve}}
Given a tool list $T$, query $q$, and a prefix of the solution path, the LLM is asked to generate the next tool name $tool_{i+1}^{pred}$ to call, and then the evaluator judges whether the name is the same as the golden answer $tool_{i+1}^{gt}$, achieving score $1$ if the same and $0$ otherwise. As mentioned in Sec. \ref{app: Reason Detail}, the LLM needs to generate a dictionary containing the next thought, as well as the next tool name and its corresponding parameters in the JSON format, while in the string format, the golden answer's thought is given, and the LLM only needs to generate the next tool name in a single line.

\subsection{\textsc{Understand}}
Given a tool list $T$, query $q$, and a prefix of the solution path, the LLM is tasked with generating the parameters $args_{i+1}^{pred}$ to call the next tool, and the score is the similarity between $args_{i+1}^{pred}$ and $args_{i+1}^{gt}$ calculated by Sentence-BERT (the same as planning evaluator). As mentioned in Sec. \ref{app: Reason Detail}, the LLM needs to generate a dictionary containing the next thought, as well as the next tool name and its corresponding parameters in the JSON format, while in the string format, the golden answer's thought and tool name are given, and the LLM needs to generate the parameters in a single line.

\subsection{\textsc{Review}}
Given a thought $t_i$ and a tool response $o_i$, the LLM is required to evaluate the tool response. It must select one of the following categories to classify the response: \textit{Success}, \textit{Internal Error}, \textit{Input Error}, \textit{Irrelevant Response}, or \textit{Unable to Accomplish}. The evaluation is scored as $1$ for a correct classification and $0$ if the classification is incorrect.

\section{API Documentation}

We manually curate extensive API documentation for each tool, following the annotation format defined by OpenAI. Compared to official RapidAPI documentation, our descriptions are more extensive and detailed, which facilitates the understanding of various tools and circumvents the failure cases due to the incomplete API documentation provided in the benchmark. Here, we provide the API documentation for BINGMap as an example in Fig. \ref{fig: API document BINGMap}. For more detailed API documentation please refer to our official benchmark code\footnote{\href{https://github.com/open-compass/T-Eval}{https://github.com/open-compass/T-Eval}}.
\newtcolorbox{AIbox}[2][]{aibox, title=#2,#1}
\begin{figure*}[!ht] 
\vspace{-5mm}
\begin{AIbox}{B.1.1 Query Generation}
{\color{blue}\bf Prompt:} \\
{
You will be given a tool list with a few API descriptions. Please carefully read the API documentation. Your task involves creating 3 varied, innovative, and concrete queries. Respond strictly with JSON. The JSON should be compatible with the TypeScript type \texttt{Response} from the following:
\begin{lstlisting}[style=prompt]
interface Response {
    // 3 generated responses based on the given tools
    cases: [
        0: {
            // list of tools selected, in the format "tool_name.api_name" e.g., AirbnbSearch.search_property_by_place
            tools: list;
            // describes one specific role, it should be a common role in our daily life
            role: string;
            // describes the detailed query content
            query: string;
        };
        // rest 2 use cases
        ...
    ];
}
\end{lstlisting}
Rule:\\
1. API name is strictly prohibited from appearing in the generated query.\\
2. Each query must use all tools: \{tool\_names\}. Query that only calls one tool will NOT be accepted.\\
3. The QUERY and ROLE must be totally different from your previous response.\\
4. The query should contain every detail, especially parameter inputs required by the APIs, so that it can be completed **without** using any external information.\\
5. The information provided in the query MUST truly exist, especially unique IDs required in APIs.\\
6. The maximum requested number of items should be limited to 3.
}
\end{AIbox} 
\caption{An example prompt of query generation.}
\label{fig: Query Generation}
\vspace{-5mm}
\end{figure*}

\begin{figure*}[!ht] 
\vspace{-5mm}
\begin{AIbox}{B.1.2 Query Refinement}
{\color{blue}\bf Prompt:} \\
{
You will be provided with 3 queries. Please carefully review and update the query so that the query can be successfully executed. Respond strictly with JSON. The JSON should be compatible with the TypeScript type \texttt{Response} from the following:
\{response\_format\}\\
Rules:\\
1. API name (e.g., search\_property\_by\_coordinates) is strictly prohibited from appearing in the query.\\
2. Provide the exact and real information in the query, Do NOT provide template information, e.g., YOUR\_FILE\_PATH.\\
3. Avoid saying `a specific xxx', `the first/second xxx' is preferred.\\
4. Fake information is strictly prohibited in the query (e.g., 1234567). You can modify part of the query so that the desired information can be obtained by other APIs to avoid generating these fake information.
}
\end{AIbox} 
\caption{An example prompt of query refinement.}
\label{fig: query refinement}
\vspace{-5mm}
\end{figure*}

\begin{figure*}[!ht] 
\vspace{-6mm}
\begin{AIbox}{B.1.3 Multi-Agent Annotation Prompt}
{\color{blue}\bf System Prompt:} \\
{
Answer the following questions as best you can. Specifically, you have access to the following APIs: \{func\_list\}.
Respond strictly with JSON. The JSON should be compatible with the TypeScript type \texttt{Response} from the following:
\begin{lstlisting}[style=prompt]
interface Response {
    // task id of the action
    id: int;
    // name of the action, must be function name
    name: string;
    // dependency/prerequisite of current action, list of task id
    dep: list;
    // input params required by current action.
    args: Record<string, any>;
    // the exact goal of executing this action
    goal: string;
}
\end{lstlisting}
Remember: \\
1. ONLY generate one action at each time.\\
2. If you have finished ALL tasks requested by the query, please reply: \{finish\_example\}\\
Begin!
}\\
{\color{orange}\bf Reviewer Prompt:} \\
{
You are an expert in discriminating if the task is accomplished. You will be provided with the following information:
\begin{lstlisting}[style=prompt]
Goal: the goal of the current task
Action: tool name of the current task
Response: response from the called tool
\end{lstlisting}
Respond strictly with JSON. The JSON should be compatible with the TypeScript type \texttt{Response} from the following:
\begin{lstlisting}[style=prompt]
interface Response {
    // explain why the task is accomplished/unaccomplished
    thought: string;
    // whether the task is done
    is_finished: boolean;
}
\end{lstlisting}
Begin!
}
\end{AIbox} 
\caption{An example prompt of Multi-Agent Annotation Prompt.}
\label{fig: multi-agent annotation prompt}
\end{figure*}

\begin{figure*}[!ht] 
\begin{AIbox}{B.2.1 Dataset Demonstration -- \textsc{Instruct}}
{\color{blue}\bf System:} \\
{
You have access to the following API:
\begin{lstlisting}[style=prompt]
{
    'name': 'AirbnbSearch.search_property_by_place', 
    'description': 'This function takes various parameters to search properties on Airbnb.', 
    'required_parameters': [{'name': 'place', 'type': 'STRING', 'description': 'The name of the destination.'}], 
    'optional_parameters': [], 
    'return_data': [{'name': 'property', 'description': 'a list of at most 3 properties, containing id, name, and address.'}]
}
\end{lstlisting}
Please generate the response in the following format:
\begin{lstlisting}[style=prompt]
{
    goal: goal to call this action
    name: API name to call
    args: JSON format API args in ONLY one line
}
\end{lstlisting}
}
{\color{orange}\bf User:} \\
{Call the function AirbnbSearch.search\_property\_by\_place with the parameter as follows: `place' is `Berlin'; }
\end{AIbox}
\caption{An example prompt in the \textsc{Instruct} dataset.}
\label{fig: dataset demonstration instruct}
\vspace{-7mm}
\end{figure*}

\begin{figure*}[!ht] 
\begin{AIbox}{B.2.2 Dataset Demonstration -- \textsc{Plan}}
{\color{blue}\bf System:} \\
{
You have access to the following API:
\begin{lstlisting}[style=prompt]
{API_docs}
\end{lstlisting}
Please generate a plan for answering the user's questions, which should be a list of actions with the following format:
\begin{lstlisting}[style=prompt]
[{
    // id of the action
    "id": number;
    // the name of the action
    "name": string;
    // input params required by this action
    "args": str(Record<string, any>);
}, ...
]
\end{lstlisting}
You can imagine args when you plan the action, and these instructions will be executed sequentially.
For example, if you want to call api1 with arg1 and arg2, you can write the following plan:
\begin{lstlisting}[style=prompt]
[
    {
        "id": 0,
        "name": "api1",
        "args": "{'arg1': 'value1', 'arg2': 'value2', ...}",
    }, ...
]
\end{lstlisting}
The args should be a dictionary in string format.
PLEASE use `' in the args dictionary and use "" in other places, DO NOT print args with value None or null.

You should only generate a list in JSON format. The list should be the full planning list without `...'
DO NOT generate any text to explain the JSON.
}

{\color{orange}\bf User:} \\
{
As a researcher studying sustainable energy technologies, I need to find properties in Berlin and review no more than three of these properties. Moreover, I need to find articles on Arxiv related to `solar energy' and get the meta-information for up to three of these articles.
}
\end{AIbox} 
\caption{An example prompt in the \textsc{Plan} dataset.}
\label{fig: dataset demonstration plan}
\end{figure*}

\begin{figure*}[!ht]
\begin{AIbox}{B.2.3 Dataset Demonstration -- \textsc{Reason}}
{\color{blue}\bf System:} \\
{
You are an assistant who can utilize external tools. You can call the following tools:
\begin{lstlisting}[style=prompt]
{API_docs}
\end{lstlisting}
If you already know the answer, please call the FinishAction to provide the final response to the answer.
}\\
{\color{orange}\bf User:} \\
{
As a researcher studying sustainable energy technologies, I need to find properties in Berlin and review no more than three of these properties. Moreover, I need to find articles on Arxiv related to `solar energy' and get the meta-information for up to three of these articles.
}\\
{\color{orange}\bf User:} \\
{
What is your thought at the current step?
}
\end{AIbox}
\caption{An example prompt in the \textsc{Reason} dataset.}
\label{fig: dataset demonstration reason}
\end{figure*}

\begin{figure*}[!ht]
\begin{AIbox}{B.2.4 Dataset Demonstration -- \textsc{Retrieve}}
{\color{blue}\bf System:} \\
{
You are an assistant who can utilize external tools. You can call the following tools: 
\begin{lstlisting}[style=prompt]
{API_docs}
\end{lstlisting}
If you already know the answer, please call the FinishAction to provide the final response to the answer.
}\\
{\color{orange}\bf User:} \\
{
As a researcher studying sustainable energy technologies, I need to find properties in Berlin and review no more than three of these properties. Moreover, I need to find articles on Arxiv related to `solar energy' and get the meta-information for up to three of these articles.
}\\
{\color{orange}\bf User:} \\
{
What is your thought at the current step?
}\\
{\color{teal}\bf Assistant:} \\
{
Find properties in Berlin.
}\\
{\color{orange}\bf User:} \\
{
What is the tool name to call at the current step?
}
\end{AIbox} 
\caption{An example prompt in the \textsc{Retrieve} dataset.}
\label{fig: dataset demonstration retrieve}
\end{figure*}

\begin{figure*}[!ht]
\begin{AIbox}{B.2.5 Dataset Demonstration -- \textsc{Understand}}
{\color{blue}\bf System:} \\
{
You are an assistant who can utilize external tools. You can call the following tools: 
\begin{lstlisting}[style=prompt]
{API_docs}
\end{lstlisting}
If you already know the answer, please call the FinishAction to provide the final response to the answer.
}\\
{\color{orange}\bf User:} \\
{
As a researcher studying sustainable energy technologies, I need to find properties in Berlin and review no more than three of these properties. Moreover, I need to find articles on Arxiv related to `solar energy' and get the meta-information for up to three of these articles.
}\\
{\color{orange}\bf User:} \\
{
What is your thought at the current step?
}\\
{\color{teal}\bf Assistant:} \\
{
Find properties in Berlin.
}\\
{\color{orange}\bf User:} \\
{
What is the tool name to call at the current step?
}\\
{\color{teal}\bf Assistant:} \\
{
AirbnbSearch.search\_property\_by\_place
}\\
{\color{orange}\bf User:} \\
{
What is the value of 'place' required by the current tool?
}
\end{AIbox}
\caption{An example prompt in the \textsc{Understand} dataset.}
\label{fig: dataset demonstration understand}
\end{figure*}

\begin{figure*}[!ht] 
\vspace{-5mm}
\begin{AIbox}{B.2.6 Dataset Demonstration -- \textsc{Review}}
{\color{blue}\bf System:} \\
{
You are presented with information about a task and its corresponding responses in the following format:
\begin{lstlisting}[style=prompt]
Goal: [The intended goal or objective of the current task.]
Name: [The name of the tool used for the task.]
Args: [The arguments or parameters supplied to the tool.]
Response: [The response or output received from the tool after execution.]
\end{lstlisting}

Based on this information, your task is to evaluate whether the goal has been achieved. Select the most appropriate option from the choices below to describe the task's outcome:

A: Success -- The task has been completed successfully and the goal is achieved.
B: Internal error -- There is a malfunction or connectivity issue within the tool itself, leading to failure.
C: Input Error -- The tool is mismatched to the query, or the parameters or arguments provided to the tool are incorrect or inadequate for the task.
D: Irrelevant Response -- The output from the tool does not align with the expected response as per the tool's description or is ambiguous.
E: Unable to Accomplish -- The tool's response indicates that the task is impossible to accomplish or the tool's response is empty.

Your output should follow the following format:
\begin{lstlisting}[style=prompt]
Answer: [Insert your choice here, choosing from A, B, C, D, and E. This should be a single character.]
\end{lstlisting}

Note that the place, the date, and the id in the parameters and the response are correct; please do not judge the correctness of the parameters and the response based on the place, the date, and id.
}\\
{\color{orange}\bf User:} \\
\begin{lstlisting}[style=prompt]
Goal: prints the details of the movie 'The Battle at Lake Changjin'\\
Name: FilmDouban.print\_detail\\
Args: {'film\_name': 'Inception'}\\
Response: {'text': 'Can not find the movie named Inception'}\\
\end{lstlisting}
\end{AIbox} 
\caption{An example prompt in the \textsc{Review} dataset.}
\label{fig: dataset demonstration review}
\end{figure*}

\begin{figure*}[!ht] 
\begin{AIbox}{D. API Documentation -- BINGMap}
\begin{lstlisting}[style=prompt]
tool_description = dict(
    name='BINGMap',
    standardized_name='bing_map',
    tool_description="Plugin for lookup map information in America",
    category="Life",
    api_list = [
        dict(
            name="get_distance",
            description="Get the distance between two locations in km.",
            required_parameters=[dict(name='start', type='STRING', description='The start location.'),
                dict(name='end', type='STRING', description='The end location.')],
            optional_parameters=[],
            return_data=[dict(name="distance", description="the distance in km.")]
        ),
        dict(
            name="get_route",
            description="Get the route between two locations in km.",
            required_parameters=[dict(name='start', type='STRING', description='The start location.'),
                dict(name='end', type='STRING', description='The end location.')],
            optional_parameters=[],
            return_data=[dict(name="route", description="the route, a list of actions.")]
        ),
        dict(
            name="get_coordinates",
            description="Get the coordinates of a location.",
            required_parameters=[dict(name='location', type='STRING', description='the location need to get coordinates.')],
            optional_parameters=[],
            return_data=[
                dict(name="latitude", description="the latitude of the location."), 
                dict(name="longitude", description="the longitude of the location.")
            ]
        ),
        dict(
            name="search_nearby",
            description = "Search for places nearby a location, within a given radius, and return the results into a list. Put the location name at the end of the query.",
            required_parameters=[dict(name='search_term', type='STRING', description='the place name')],
            optional_parameters=[
                dict(name='places', type='STRING', description='the name of the location.'),
                dict(name='latitude', type='FLOAT', description='the latitude of the location.'),
                dict(name='longitude', type='FLOAT', description='the longitude of the location.'),
                dict(name='radius', type='NUMBER', description='radius in meters.')
            ],
            return_data=[
                dict(name="places", description="the list of places, each place is a dict with name and address, at most 5 places.")
            ]
        ),
    ]
)
\end{lstlisting}
\end{AIbox} 
\caption{An example API document: BINGMap.}
\label{fig: API document BINGMap}
\end{figure*}

%% file: tables/new_dataset.tex
\begin{table}[!ht]
\small
\caption{The statistics of the evaluation datasets in \shortname.}
\label{tab:data_stat}
\centering
\begin{tabular}{c| c }
    \toprule
        \textbf{Dataset} & \textbf{Test Cases} \\
    \midrule
    \textsc{Instruct} & 2660\\
    \textsc{Retrieve} & 6426\\
    \textsc{Plan} & 553\\
    \textsc{Reason} & 6426\\
    \textsc{Review} & 487\\
    \textsc{Understand} & 6753\\
    \midrule
    Total & 23305\\
    \bottomrule
\end{tabular}
\end{table}